\documentclass{INTERSPEECH2023}

\interspeechcameraready 

\usepackage{booktabs}
\usepackage{subcaption}
\usepackage{url}

\title{Analyzing Multiple-Choice Reading and Listening Comprehension Tests}
\name{Vatsal Raina, Adian Liusie, Mark Gales \thanks{This research is funded by the EPSRC (The Engineering and Physical Sciences Research Council) Doctoral Training Partnership (DTP) PhD studentship and supported by Cambridge Assessment, University of Cambridge and ALTA.}}
\address{
  ALTA Institute/Department of Engineering, Cambridge University}
\email{\{vr311,al826,mjfg\}@cam.ac.uk}

\begin{document}

\maketitle
 
\begin{abstract}

Multiple-choice reading and listening comprehension tests are an important part of language assessment. Content creators for standard educational tests need to carefully curate questions that assess the comprehension abilities of candidates taking the tests. However, recent work has shown that a large number of questions in general multiple-choice reading comprehension datasets can be answered without comprehension, by leveraging world knowledge instead. This work investigates how much of a contextual passage needs to be read in multiple-choice reading based on conversation transcriptions and listening comprehension tests to be able to work out the correct answer. We find that automated reading comprehension systems can perform significantly better than random with partial or even no access to the context passage. These findings offer an approach for content creators to automatically capture the trade-off between comprehension and world knowledge required for their proposed questions.

\end{abstract}

\section{Introduction}

Multiple-choice reading and listening comprehension tests serve as essential tools for evaluating language proficiency in educational settings \cite{AldersonJ.Charles.2000AR}. In particular, multiple-choice questions permit fast and automated objective assessment of candidates' abilities. The creation of these standardized tests necessitates the careful selection of questions that accurately assess candidates' comprehension abilities. It is of interest for content creators to develop a framework to categorize the quality of questions used in assessment across several criteria such as complexity and diversity \cite{raina2022multiple}.   

\begin{figure}[t!]
    \centering
        \includegraphics[width=1.0\linewidth]{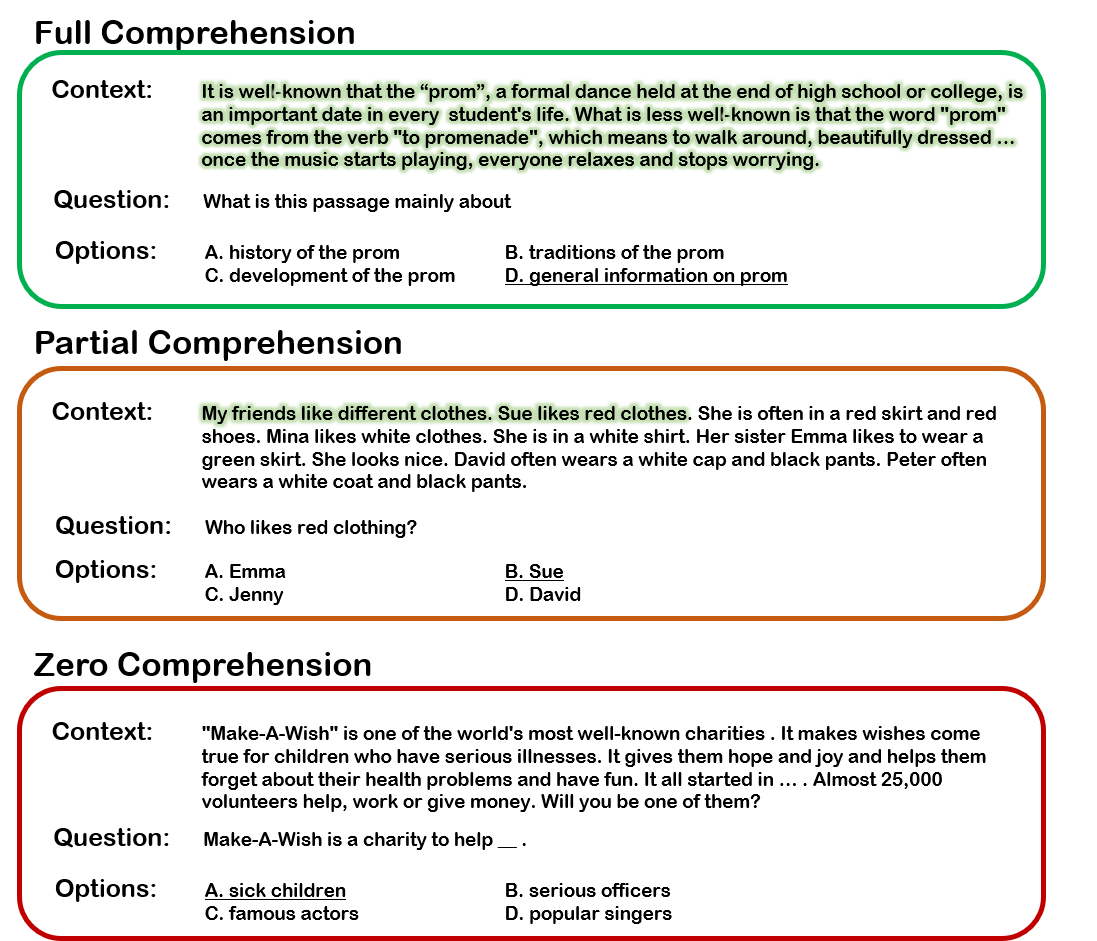}
    \caption{Example questions that can be answered with full, partial and zero comprehension respectively.}
    \label{fig:example}
\end{figure}

However, recent work \cite{liusie-etal-2023-world} has identified an issue within general multiple-choice reading comprehension datasets sourced from real tests — many questions can be answered correctly without language learners truly comprehending the passage, merely by relying on prior world knowledge. This work builds upon the concept of world knowledge in reading comprehension and aims to explore the extent to which contextual passages must be read/heard in multiple-choice reading/listening tests based on conversation transcriptions and listening comprehension assessments to deduce the correct answer. For example, a candidate may be able to deduce the correct answer to a large number of the comprehension questions by only reading the first sentence. Typically language learners may not understand the whole context and only partially comprehend the sentences. Figure \ref{fig:example} demonstrates three multiple-choice questions with varying degrees of required comprehension. Full comprehension, when the whole passage must be read in order to determine the correct answer. Partial comprehension, when the correct answer can be deduced from reading only a small part of the context. Finally, zero comprehension in the extreme case where the correct answer can be deduced without reading the context at all and by using world knowledge instead. For instance, in the zero comprehension example in Figure \ref{fig:example}, without any need to read the context it is obvious that the answer is \textit{sick children} as the question asks about charities.

Information about the extent of comprehension required in reading and listening tests can act as a core component in the question assessment framework \cite{raina2022multiple, liusie2023camchoice}. The degree of comprehension required can vary across the nature of the comprehension dataset. In this work, we consider a range of publicly available datasets that are very different in nature including commonsense-based reasoning, logical reasoning and multi-turn dialogue, speech transcriptions. We make the following contributions in this work:
\begin{itemize}
    \item Portability of world knowledge and partial comprehension systems from standard multiple-choice reading comprehension to dialogue and speech.
    \item A thorough investigation of the degree of partial comprehension from zero comprehension (world knowledge) to full comprehension.
\end{itemize}
We emphasize the need for content creators to carefully and explicitly consider the extent of comprehension required for the questions they generate in order to better capture how language learners may interact with the deployed questions in tests.


\section{Related work}

\cite{liusie-etal-2023-world} indicates world knowledge is prevalent in several standard multiple-choice reading comprehension systems, reinforcing whether machine reading comprehension systems fully leverage the context for the desired comprehension task \cite{sugawara2020assessing, kaushik2018much, jia2017adversarial, si2019does}. \cite{liusie-etal-2023-world} further introduces two performance metrics, effective number of options and mutual information of the context, to assess the extent to which world knowledge is used in these reading comprehension systems. We extend the work on world knowledge to investigate the spectrum between zero comprehension to full comprehension of real multiple-choice comprehension questions for text-based, dialogue-based and speech-based contexts.

Previous work investigated automated approaches to assess the quality of comprehension questions. \cite{raina2022multiple} present a framework to assess the quality of generated multiple-choice questions for comprehension. Four main qualities are identified: grammatical fluidity, answerability, diversity and complexity. Our work on assessing the extent to which the context needs to be read acts as an extension to this framework to capture the comprehensibility of the generated questions. 

Due to the lack of appropriately annotated speech corpora, several works investigate porting text-based systems for listening comprehension tasks. \cite{chung2018supervised} explores applying a text-based question answering system on the TOEFL listening comprehension multiple-choice test from \cite{tseng2016towards}. \cite{raina2021initial} further investigates the transfer learning style approach for extractive comprehension from SQuAD 2.0 \cite{Rajpurkar2018KnowWY} to a proprietary spoken question answering task, with a particular focus on the impact of automatic speech recognition (ASR) errors. Our approach ports systems from a multiple-choice reading comprehension task to a multiple-choice listening comprehension task to identify the extent to which comprehension of the context is required.

\section{Multiple-choice comprehension}

\subsection{Task}

Multiple-choice comprehension is a common assessment technique to assess the comprehension abilities of candidates in standardized tests \cite{frizelle_o'neill_bishop_2017}. Given a context passage, $C$ and a question, $Q$, the correct answer must be deduced from a discrete set of $N$ answer options, $\{O\}$. Hence, it is required to deduce the correct answer by comprehending the question and using the context passage as the information source to identify which answer option is the most suitable.


\subsection{Machine comprehension}

\begin{figure}[t!]
    \centering
        \includegraphics[width=1.0\linewidth]{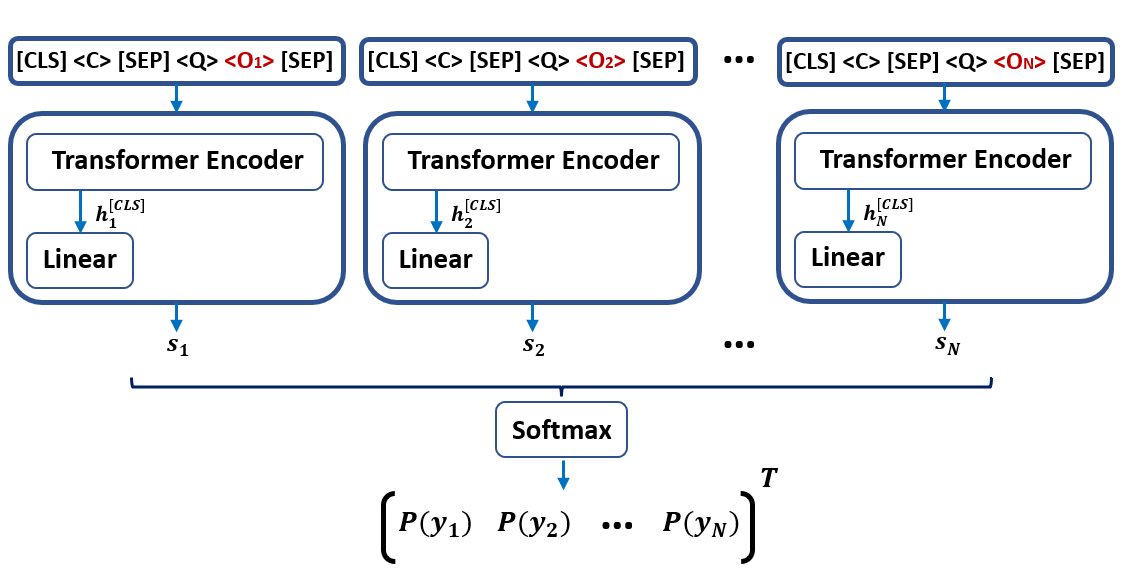}
    \caption{The architecture for multiple-choice machine comprehension with context, $C$, question, $Q$ and $N$ options, $\{O\}$.}
    \label{fig:arch}
\end{figure}

Machine comprehension performs the comprehension task using automated systems. Machine reading and listening comprehension for multiple-choice tests is a well researched area with state-of-the-art systems \cite{Zhang2021RetrospectiveRF, Yamada2020LUKEDC, Zaheer2020BigBT, wang2021logicdriven} competing and out-performing humans on public benchmarks \cite{Clark2018ThinkYH, Lai2017RACELR, Trischler2017NewsQAAM, Yang2018HotpotQAAD}. 

In this work, the machine comprehension system's architecture replicates the standard multiple-choice machine reading comprehension systems from \cite{Yu2020ReClorAR, raina-gales-2022-answer} and depicted in Figure \ref{fig:arch}. Each option is separately encoded with the question and the context to generate a score. A softmax layer converts the scores associated with each option into a probability distribution where at inference time the predicted answer is taken to be the option with the greatest probability. The parameters of the core transformer \cite{vaswani2017attention} encoder and the linear layer are shared across all options. Hence, there is no requirement for the number of options at training and inference time to match.

\subsection{World knowledge}

It is expected that information must be used from both the context passage and the question to determine the correct answer. If the answer can be deduced without the context, it suggests `world knowledge' \cite{liusie-etal-2023-world} is sufficient to answer the question. We train a context-free system where the context is omitted to determine the extent to which world knowledge can be leveraged for comprehension. Table \ref{tab:systems} summarizes the main differences between the standard and context-free systems where [CLS] and [SEP] denote classification and separation tokens respectively.

\begin{table}[htbp!]
\centering
\caption{Format for multiple-choice comprehension systems.}
\begin{small}
    \begin{tabular}{ll}
    \toprule
System &  Format \\
\midrule
Standard & \texttt{[CLS]}$<$$C$$>$\texttt{[SEP]}$<$$Q$$>$$<$$O_i$$>$\texttt{[SEP]} \\
Context-free & \texttt{[CLS]}$<$$Q$$>$$<$$O_i$$>$\texttt{[SEP]} \\
   \bottomrule
    \end{tabular}
    \end{small}
    \label{tab:systems}
\end{table}

\subsection{Partial context}

Language learners often can shortcut reading the whole context passage in comprehension tasks and still correctly answer the question. Hence, we devise a simple approach to investigate the extent to which a context must be comprehended in order to determine the correct answer to standard multiple-choice questions. A standard system (see Table \ref{tab:systems}) trained with the full context is taken and applied at inference time to questions with only partial access to the context. After applying tokenization of the context, only $\tau$\% of the context tokens are retained and input to the standard system. $\tau$ can be varied to determine how much of the context is necessary for comprehension.

\section{Experiments}

\subsection{Data}
\label{sec:data}

Several multiple-choice reading/listening comprehension datasets are used in this work including: RACE++ \cite{pmlr-v101-liang19a}, ReClor \cite{Yu2020ReClorAR}, COSMOSQA \cite{Huang2019CosmosQM}, DREAM \cite{sun2019dream} and IBM-Debater \cite{mirkin-etal-2018-listening}.

\begin{table}[htbp!]
\centering
\caption{Dataset statistics. Relevant examples are underlined.}
\begin{small}
    \begin{tabular}{l|rrr|c}
    \toprule
& \texttt{TRN} & \texttt{DEV} & \texttt{EVL} & \#options \\
\midrule
RACE++ & \underline{100,388} & \underline{5,599} & \underline{5,642} & 4 \\
COSMOSQA & 25,262 & \underline{2,985} & -- & 4 \\
ReClor & 4,638 & \underline{500} & 1000 & 4 \\
DREAM & 6,116 & 2,040 & \underline{2,041} & 3\\
IBM-Debater & -- & -- & \underline{200} & 2 \\
   \bottomrule
    \end{tabular}
    \end{small}
    \label{tab:alldata}
\end{table}

\noindent \textbf{RACE++} is a dataset of English reading comprehension questions for Chinese high school students. The questions are collected at three levels: middle school, high school and college level, corresponding to increasing levels of complexity.
\newline\newline
\noindent \textbf{COSMOSQA} is a large scale commonsense-based reading comprehension dataset with four options per question. For this work, 2,985 examples from the development set is used.
\newline\newline
\noindent \textbf{ReClor} is a logical reasoning dataset at a graduate student
level with four options per question. This is a challenging dataset as graduate students achieve an accuracy of 63\%. 500 examples from the development split are used for this work (the test set is hidden).
\newline\newline
\noindent \textbf{DREAM} is a multiple-choice (three options) reading comprehension dataset that focuses on dialogue understanding. These dialogue are multi-turn and multi-party. It contains 10,197 questions and 6,444 dialogues, which were collected from English-as-a-foreign-language examinations. This work uses the 2,041 questions from the test split. The context is constructed by concatenating all dialogues into a single text.
\newline\newline
\noindent \textbf{IBM-Debater} consists of 200 spontaneous speeches arguing for or against 50 controversial topics. The dataset is structured to form a multiple-choice listening comprehension task by formulating each speech as a question that is aimed at confirming or rejecting the argument in a speech. Hence, each question has a binary class label with the transcribed speech acting as the context. The transcriptions are available as both manual and automatic speech recognition transcriptions.

\subsection{Training details and hyperparameters}

Two systems are trained on the large RACE++ training dataset (see Table \ref{tab:systems}): 1. A standard multiple-choice reading comprehension system with access to the context; 2. A context-free system without access to the context. Both systems are deep ensembles of 3 models that specifically use the large~\footnote{Model configuration at: \url{https://huggingface.co/google/electra-large discriminator/blob/main/config.json}} ELECTRA \cite{clark2020electra} pre-trained language model in the form of the multiple-choice machine comprehension architecture of Figure \ref{fig:arch}.

Each model has 340M parameters.
Grid search was performed for hyperparameter tuning of the standard system with the initial setting of the hyperparameter values by the systems from \cite{raina-gales-2022-answer}. Apart from the default values used for various hyperparameters, the grid search was performed for the maximum number of epochs $\in \{2,5,10\}$; learning rate $\in \{2e-7, 2e-6, 2e-5\}$; batch size $\in \{2,4\}$. Training was performed for 2 epochs at a learning rate of 2e-6 with a batch size of 4 and inputs truncated to 512 tokens at both training and inference time.
Cross-entropy loss was used at training time with models built using NVIDIA A100 graphical processing units with training time under 4 hours per model. The context-free system had its hyperparameters selected to be identical to the standard system.

\subsection{Assessment}

Accuracy is used as the standard performance metric for inference on all datasets. The evaluation process aims to assess two aspects of the multiple-choice questions in each dataset: 1. the ability to use world knowledge in order to determine the correct answer and consequently the effective number of options per question; 2. the extent to which the context must be read/listened to determine the correct answer. The former is assessed by comparing the accuracy of a context-free comprehension system against a standard multiple-choice comprehension system while the latter is assessed by varying the amount of context available to a standard multiple-choice reading comprehension system at test time.


\section{Results}

Multiple-choice questions are assessed for comprehensibility in terms of both world knowledge and partial access to the context.

\subsection{World knowledge}

\begin{figure*}[t!]
    \centering
    \begin{subfigure}[t]{0.32\linewidth}
        \centering
        \includegraphics[width=1.0\linewidth]{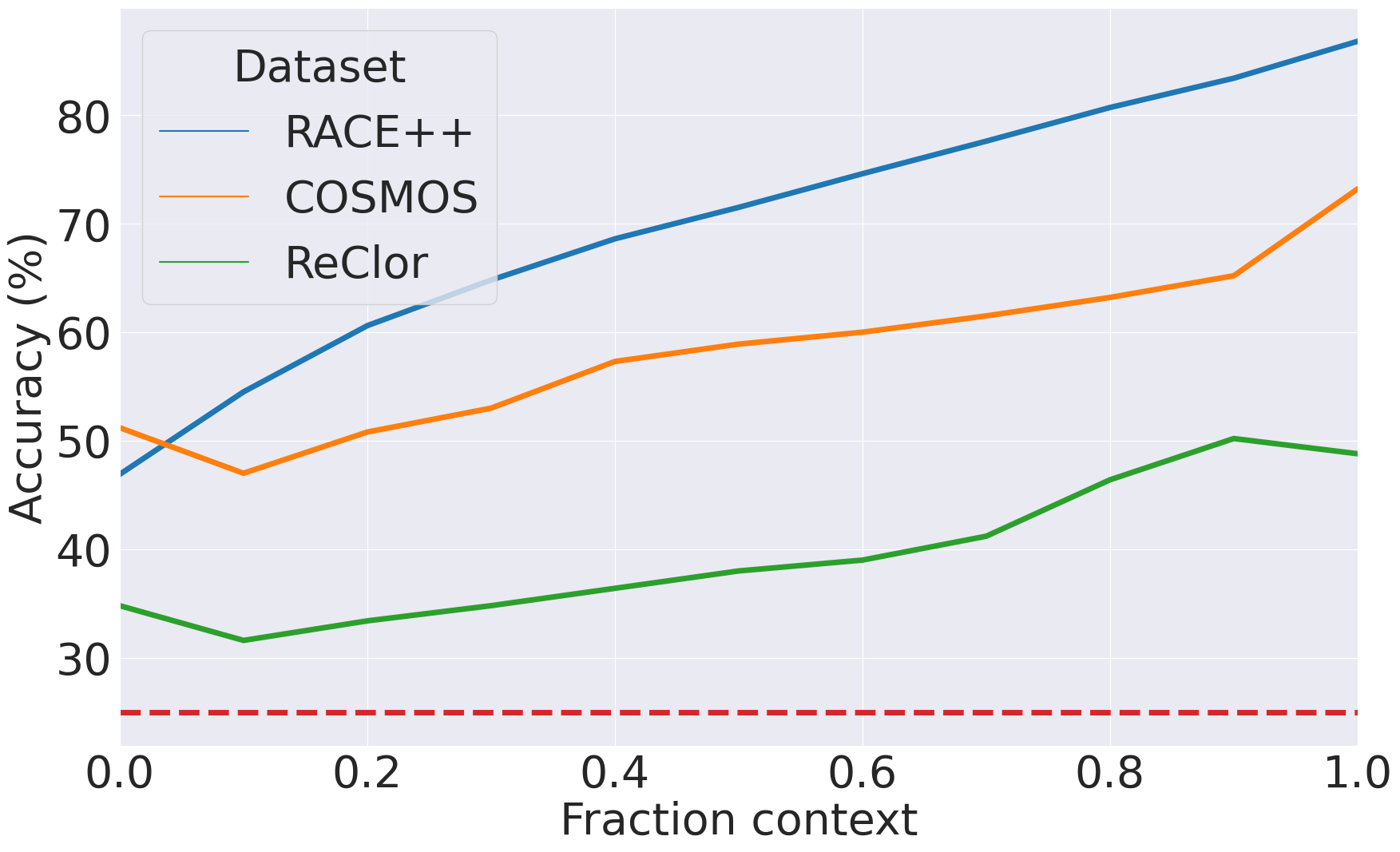}
        \caption{Text}
    \end{subfigure}
    \begin{subfigure}[t]{0.32\linewidth}
        \centering
        \includegraphics[width=1.0\linewidth]{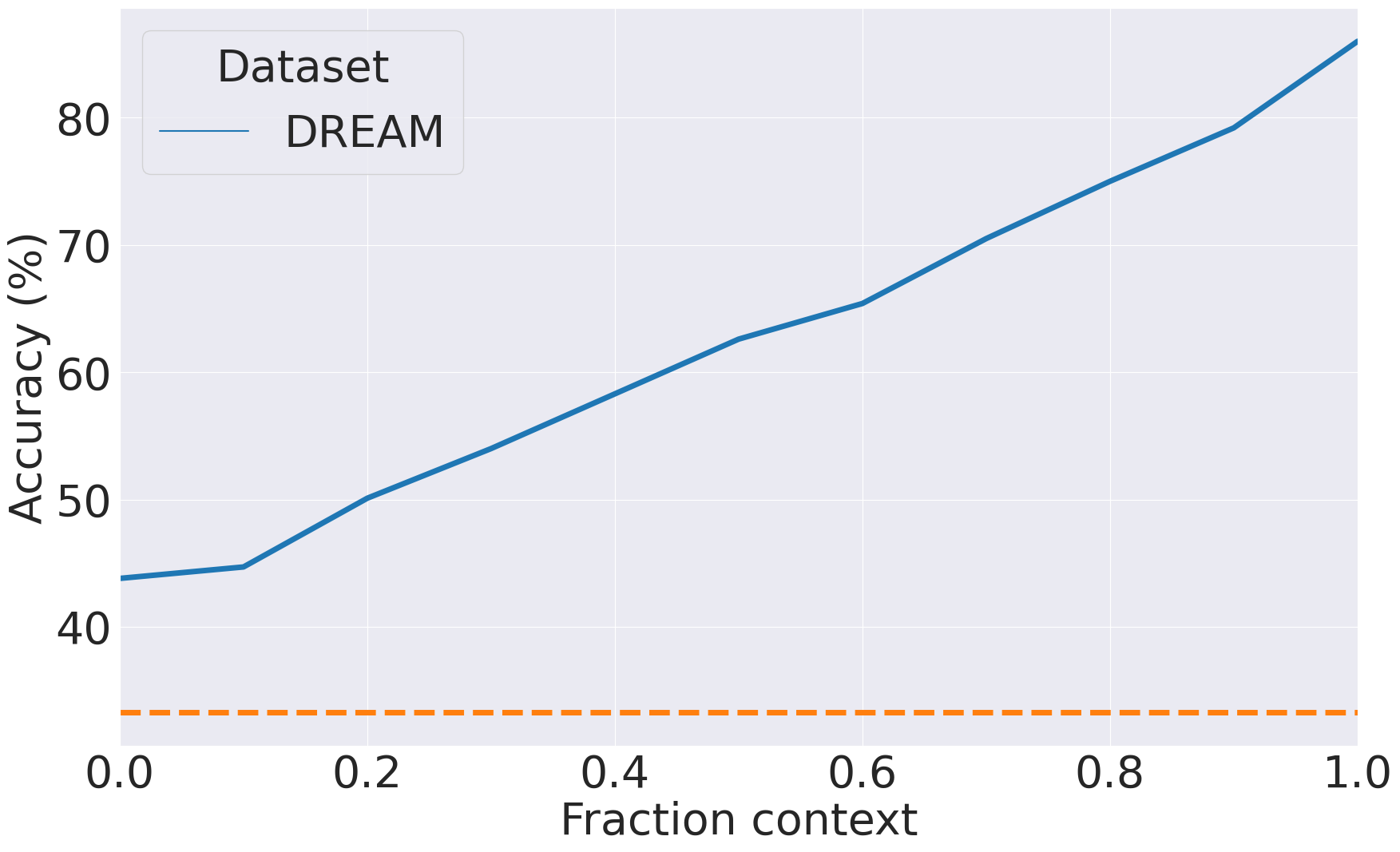}
        \caption{Dialogue}
    \end{subfigure}
    \begin{subfigure}[t]{0.32\linewidth}
        \centering
        \includegraphics[width=1.0\linewidth]{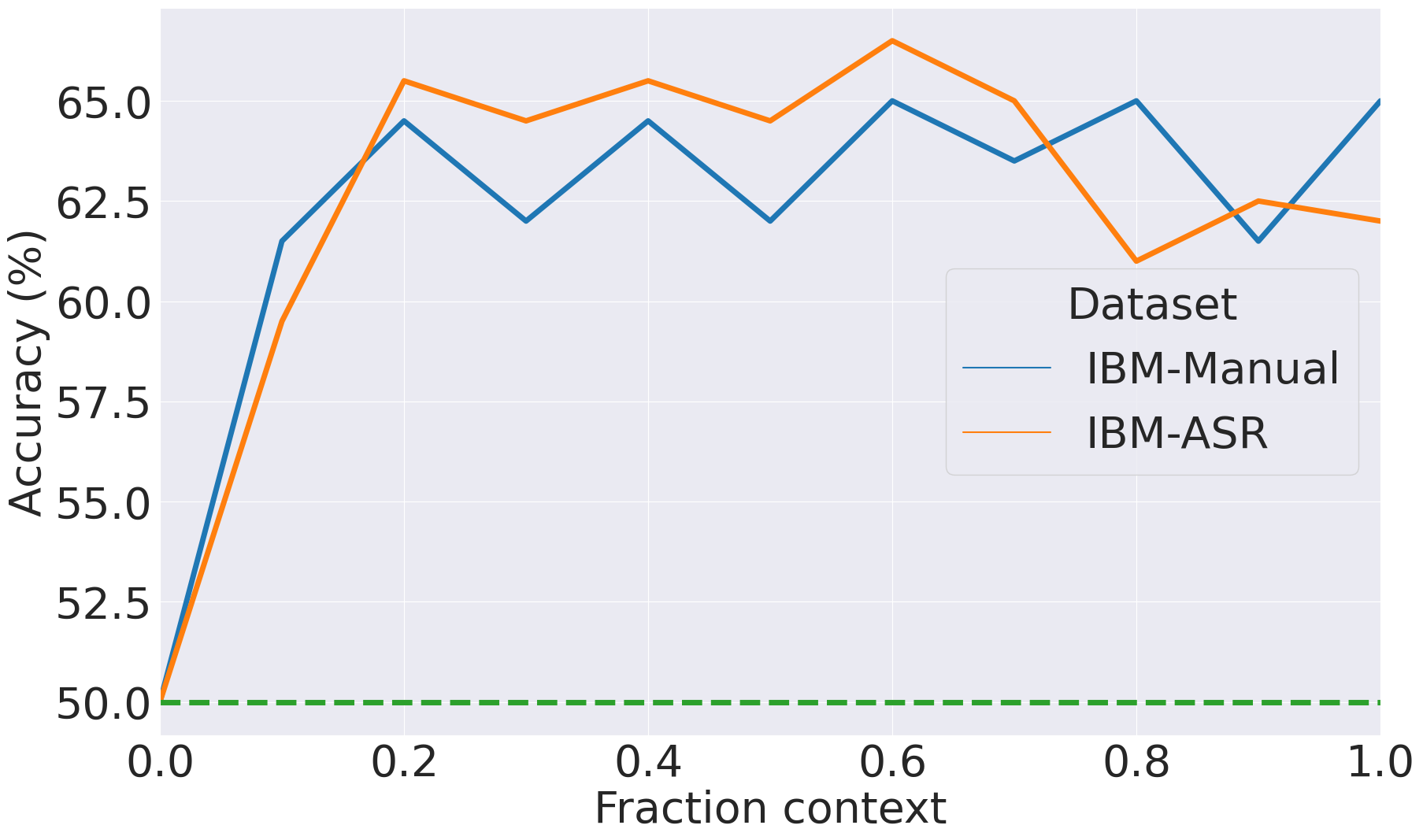}
        \caption{Speech}
    \end{subfigure}
    \caption{Accuracy with partial context access. Points are plotted at 10\% intervals.}
    \label{fig:partial_context}
\end{figure*}

Table \ref{tab:baseline} presents the prevalence of world knowledge across a range of reading and listening comprehension datasets. As both the standard and the context-free systems are trained on the RACE++ dataset, Table \ref{tab:baseline} further presents the portability of the systems to different forms of reading/listening comprehension.

\begin{table}[htbp!]
\centering
\caption{Accuracy of standard and context-free systems trained on RACE++ in-domain and out-of-domain.}
\begin{small}
    \begin{tabular}{l|ccc}
    \toprule
& Standard & Context-free & Random \\
 \midrule   
RACE++ & 86.8 & 59.1 & 25.0 \\
COSMOSQA & 73.2 & 52.8 & 25.0 \\
ReClor & 48.8 & 38.0 & 25.0 \\
DREAM & 86.0 & 46.1 & 33.3 \\
IBM-manual & 65.0 & 50.0 & 50.0 \\
IBM-ASR & 62.0 & 50.0 & 50.0 \\
   \bottomrule
    \end{tabular}
    \end{small}
    \label{tab:baseline}
\end{table}

As in \cite{liusie-etal-2023-world}, the reading comprehension datasets of RACE++, COSMOSQA and ReClor observe significant presence of world knowledge. In particular, the context-free system on RACE++ achieves an accuracy of 59.1\% despite having no access to the contextual passage that is more than double the accuracy of a random baseline. The ported context-free system also out-performs the 25\% random baseline for commonsense reasoning and logical reasoning for COSMOSQA and ReClor respectively. Note, ReClor is a more challenging reading comprehension dataset than COSMOSQA and RACE++ \cite{Yu2020ReClorAR}, confirmed by the standard RACE++ trained system getting an accuracy of 73.2\% on COSMOSQA but 48.8\% on ReClor. Systems trained directly on COSMOSQA, ReClor observe a similar pattern \cite{liusie-etal-2023-world}. 

From Table \ref{tab:baseline}, both the context-free and the standard systems port across well to dialogues in the DREAM dataset. As before, the DREAM dataset demonstrates the presence of world knowledge as the context-free system surpasses the random baseline of 33\% to achieve 46\%. It is further interesting to observe the standard system ported from RACE++ gets an accuracy of 86\%, which approaches the state-of-the-art performance of standard systems trained on DREAM \cite{zhang2022hrca+}.

However, the context-free system performs randomly on the speech transcriptions from the IBM-Debater dataset. This is an expected result as the speeches are reformulated into listening comprehension questions by posing whether the speech is pro or con a specific controversial topic (see Section \ref{sec:data}). As the speeches are balanced for each topic, it is impossible to use world knowledge for a context-free system to deduce the argument in the speech without listening to it. The standard system, with access to the speech transcription, gets an accuracy of 65\% with manual transcriptions and 62\% with ASR transcriptions, comparable to \cite{mirkin-etal-2018-listening}. Hence, the presence of ASR errors leads to a small drop in performance for binary classification.

\subsection{Partial information access}

This section investigates to what extent the context passage must be read or listened. Figure \ref{fig:partial_context} presents the accuracy with partial access to the context, varying from zero to full access, for text, dialogue and speech-based comprehension questions. 

All results are presented using the standard system trained on RACE++. Hence, the accuracy with 0\% access to the context on the plots differs in performance from the context-free system applied to the datasets from Table \ref{tab:alldata} - the context-free system's performance can expect to be an upperbound of performance with world knowledge as the system has explicitly been trained to try and deduce the correct answer without using the context. It is notable from Figure \ref{fig:partial_context} that both the text-based and dialogue based reading comprehension datasets all start above the random line while the speech-based listening comprehension dataset begins at random accuracy, agreeing with Table \ref{tab:baseline}.

Figure \ref{fig:partial_context} depicts that the text-based reading comprehension datasets increase linearly (approximately) with increasing access to the context passage. Such a linear relationship indicates that information required to deduce the correct answer is evenly distributed throughout the context passage. A similar behaviour is observed with DREAM, though the slow start indicates that information may be more disjoint in order to deduce the correct answer as emphasized in the original release of the DREAM dataset \cite{sun2019dream}. In contrast, a very different shape is observed for the speech transcriptions: there is a sharp increase on the IBM-Debater dataset with increased access to the speech and then the performance plateaus. Such a shape suggests the information is front-heavy where it is possible to deduce the side of the argument made in a speech using the first sentence. 

\begin{table}[htbp!]
\centering
\caption{Accuracy on IBM with 20\% access to the context.}
\begin{small}
    \begin{tabular}{l|cc}
    \toprule
& Manual & ASR   \\
 \midrule   
Beginning [0-20\%] & 64.5 & 65.5 \\
Random & 58.0 & 57.0 \\
End [80-100\%] & 52.5 & 55.5 \\
   \bottomrule
    \end{tabular}
    \end{small}
    \label{tab:20_fraction}
\end{table}

Table \ref{tab:20_fraction} further investigates the extent to which information is unevenly distributed in the IBM-Debater speeches. From Figure \ref{fig:partial_context}, 20\% is used as an appropriate operating point to compare the performance with access to only the beginning extract of the context against the end and random extracts.
For both the manual and the ASR transcriptions the performance is the highest for the beginning 20\% and lowest for the end 20\%, confirming the information to deduce the correct answer is concentrated at the beginning of the context.
Future work should consider evaluating how performance varies with access to the easiest vs the most difficult sentences as the easiest sections mimic the parts of the context a language learner understands \footnote{Initial experiments with sentence complexity based on standard vocabulary levels did not observe a statistically significant difference between the easiest and most difficult 20\% according to text readability.}.

Content creators are encouraged to plot similar characteristic graphs for newly proposed questions to gauge the degree of comprehension required by language learners.

\section{Conclusions}

This work highlights the trade-off between contextual comprehension and world knowledge in multiple-choice reading and listening comprehension tests. We found that automated reading comprehension systems perform significantly better than random, even with limited access to the context passage. These findings provide content creators with an approach to capture the balance between comprehension and world knowledge in their questions.
We further investigated to what extent a context needs to be read before the correct answer can be deduced, finding that it is possible to answer some questions across several reading/listening comprehension datasets with only access to a fraction of the context. Overall, our findings guide content creators in constructing more valid and reliable assessments, ensuring accurate evaluation of language proficiency.

\section{Limitations}

A limitation for the IBM-Debater dataset is that the contexts have been truncated to 512 tokens prior to any experiments despite the average length being approximately 1000 tokens to use the standard pretrained language model finetuned on RACE++.



\bibliographystyle{IEEEtran}
\bibliography{mybib}

\end{document}